\documentclass[letterpaper, 10 pt, conference]{cls/ieeeconf}
\IEEEoverridecommandlockouts
\let\labelindent\relax

\usepackage[table,usenames,dvipsnames]{xcolor}      
\usepackage{enumitem}                               
\usepackage{extarrows}                              
\usepackage[noadjust]{cite}                         

\usepackage{amsmath,amssymb,amsfonts,amsthm,dsfont} 
\usepackage{algorithm,algorithmicx,listings}        
\usepackage[noend]{algpseudocode}			        

\usepackage{graphicx,tabularx,subcaption}
\usepackage[export]{adjustbox}
\usepackage{multirow,multicol,array,rotating,diagbox}

\usepackage[font={small}]{caption}
\usepackage[titletoc,title]{appendix}

\usepackage{comment}
\usepackage{balance}
\usepackage[breaklinks=true, colorlinks, bookmarks=true, citecolor=Black, urlcolor=Violet,linkcolor=Black]{hyperref}

\def\argmin{\mathop{\arg\min}\limits}

\newcommand{\crl}[1]{\left\{#1\right\}}

\theoremstyle{definition}

\newtheorem*{assumption*}{Assumption}
\newtheorem*{problem*}{Problem}

\theoremstyle{remark}

\newtheorem*{solution*}{Solution}

\newcommand{\calD}{{\cal D}}
\newcommand{\calE}{{\cal E}}
\newcommand{\calF}{{\cal F}}

\newcommand{\calM}{{\cal M}}

\newcommand{\calP}{{\cal P}}
\newcommand{\calQ}{{\cal Q}}

\newcommand{\bfp}{\mathbf{p}}
\newcommand{\bfq}{\mathbf{q}}

\newcommand{\bfv}{\mathbf{v}}

\newcommand{\bftheta}{\boldsymbol{\theta}}

\newcommand{\bfA}{\mathbf{A}}

\newcommand{\bfC}{\mathbf{C}}
\newcommand{\bfD}{\mathbf{D}}
\newcommand{\bfE}{\mathbf{E}}

\newcommand{\bfG}{\mathbf{G}}

\newcommand{\bfI}{\mathbf{I}}

\newcommand{\bfL}{\mathbf{L}}

\newcommand{\bfR}{\mathbf{R}}

\newcommand{\bfV}{\mathbf{V}}
\newcommand{\bfW}{\mathbf{W}}

\def\thetitle{Mesh Reconstruction from Aerial Images for Outdoor Terrain Mapping Using Joint 2D-3D Learning}
\def\theauthor{Qiaojun Feng, Nikolay Atanasov}
\def\thekeywords{}

\hypersetup{
  pdfauthor={\theauthor},%
  pdftitle={\thetitle},%
  pdfsubject={\thetitle},%
  pdfkeywords={\thekeywords}
}

\IEEEoverridecommandlockouts

\title{\LARGE \bf \thetitle}

\author{Qiaojun Feng \and Nikolay Atanasov
\thanks{We gratefully acknowledge support from NSF NRI CNS-1830399.}%
\thanks{The authors are with the Department of Electrical and Computer Engineering, University of California San Diego, La Jolla, CA 92093, USA {\tt\small \{qjfeng,natanasov\}@ucsd.edu}}
}

\begin{document}
\maketitle

\begin{abstract}
This paper addresses outdoor terrain mapping using overhead images obtained from an unmanned aerial vehicle. Dense depth estimation from aerial images during flight is challenging. While feature-based localization and mapping techniques can deliver real-time odometry and sparse points reconstruction, a dense environment model is generally recovered offline with significant computation and storage. This paper develops a joint 2D-3D learning approach to reconstruct local meshes at each camera keyframe, which can be assembled into a global environment model. Each local mesh is initialized from sparse depth measurements. We associate image features with the mesh vertices through camera projection and apply graph convolution to refine the mesh vertices based on joint 2-D reprojected depth and 3-D mesh supervision. Quantitative and qualitative evaluations using real aerial images show the potential of our method to support environmental monitoring and surveillance applications.
\end{abstract}

\section{Introduction}
\label{sec:introduction}

Recent advances in sensors, processors, storage and communication devices have set the stage for mobile robot systems to significantly contribute in environmental monitoring, security and surveillance, agriculture, and many other applications. Constructing terrain maps onboard an unmanned aerial vehicle (UAV) using online sensory data would be very beneficial in these applications. With inertial measurement unit (IMU), GPS, and camera sensors, a UAV can localize itself and incrementally reconstruct the geometric structure of the traversed terrain. Near infrared cameras can additionally provide normalized difference vegetation index measurements for vegetation assessment and semantic segmentation can enrich the map. 

This paper considers the problem of building a terrain model in the form of a mesh of an outdoor environment using a sequence of overhead RGB images obtained onboard a UAV. We assume that the UAV is running a localization algorithm, based on visual-inertial odometry (VIO)~\cite{Qin2018VINS} or simultaneous localization and mapping (SLAM)~\cite{Cadena2016SLAM}
, which estimates its camera pose and the depths of a sparse set of tracked image keypoints. One approach for outdoor terrain mapping is to recover depth images at each camera view using dense stereo matching. The depth images can be fused to generate a point cloud and triangulate a mesh surface. While specialized sensors and algorithms exist for real-time dense stereo matching, they are restricted to a limited depth range, much smaller than commonly seen in aerial images. Moreover, due to the limited depth variation, the recovered point cloud might not be sufficiently dense for accurate mesh reconstruction. Recently, depth completion methods \cite{Ma2018Sparse,Chen2019Joint} using deep learning have shown promising performance on indoor \cite{Sturm2012RGBD} and outdoor datasets \cite{Geiger2013KITTI}. However, aerial images are different from the ground-level RGBD datasets used to train these algorithm. Due to the limited availability of aerial image datasets for supervision, learning-based methods have not yet been widely adopted for outdoor terrain mapping.

\begin{figure}[t]
  \centering
  \includegraphics[width=\linewidth,trim=0mm 0mm 0mm 0mm, clip]{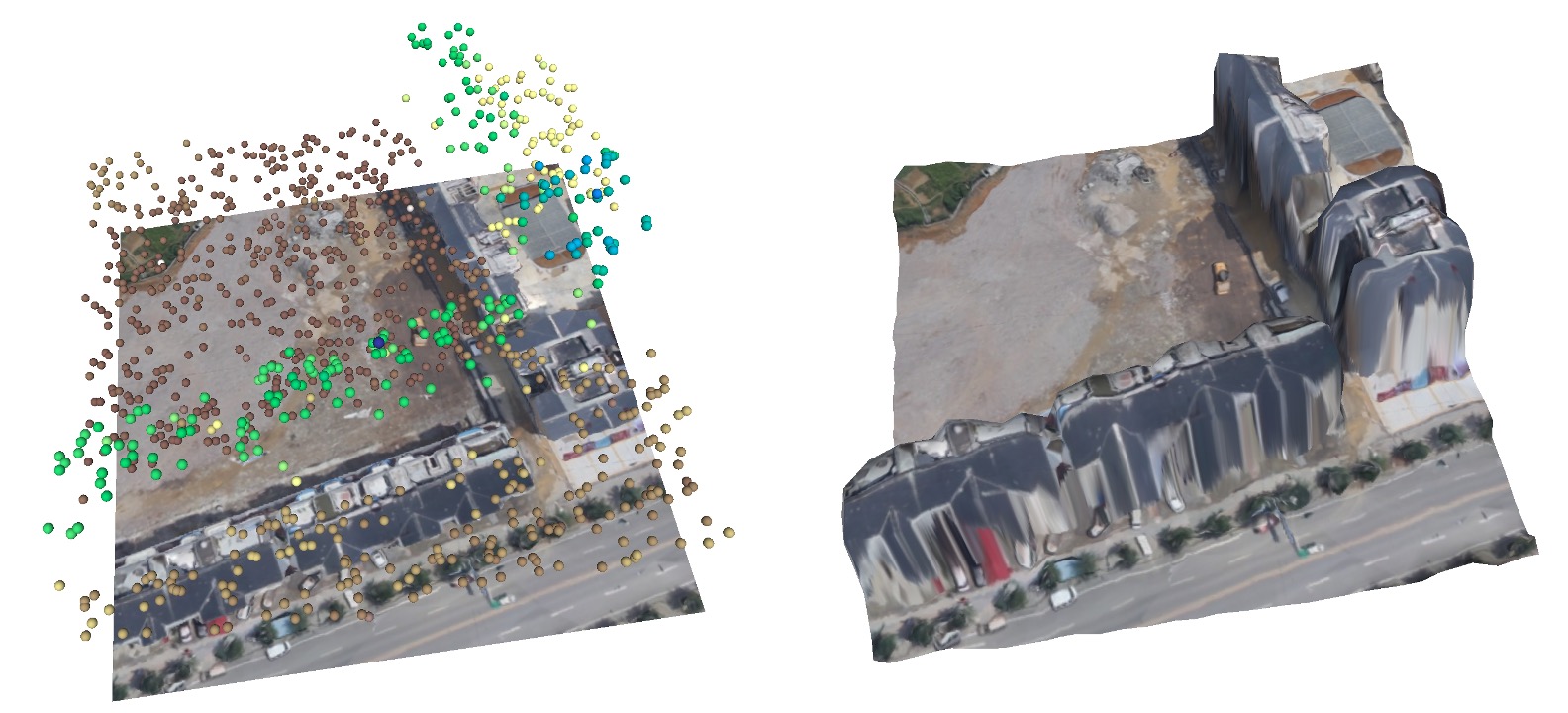}
  \caption{This paper develops a method for 3-D mesh reconstruction (right) from aerial RGB images and noisy sparse depth measurements (left) to support outdoor terrain mapping.}
\label{fig:intro}
\end{figure}

We propose a learning-based method for mesh reconstruction using a single RGB image with sparse depth measurements. Fig.~\ref{fig:intro} shows an example input and mesh reconstruction. Our main \textbf{contribution} is to show that depth completion and mesh reconstruction are closely related problems. Inspired by depth completion techniques, we propose a coarse-to-fine strategy, composed of initialization and refinement stages for mesh reconstruction. In the initialization stage, we use only the sparse depth measurements to fit a coarse mesh surface by minimizing a 2-D rendered depth loss.  In the refinement stage, we leverage both the RGB image and the sparse depth information. We extract deep convolutional image features and associate them with the vertices of the initial mesh through camera projection. The mesh is subsequently refined using a graph convolution model to predict vertex deformations that minimize a weighted 3-D geometric surface loss and 2-D rendered depth loss. Given the camera poses, we can fuse the optimized local meshes into a global mesh model of the environment. 
We build an aerial image dataset with ground-truth depth, noisy sparse depth measurements, and multiple camera trajectories based on WHU MVS/Stereo dataset~\cite{Liu2020WHU} and details can be found on \href{https://github.com/FengQiaojun/TerrainMesh}{https://github.com/FengQiaojun/TerrainMesh}.

\section{Related Work}
\label{sec:related_work}

\noindent \textbf{Depth Completion.}
Predicting depth from RGB images enables artificial perception systems to recover 3-D environment structure \cite{Godard2019Monodepth2,Gordon2019Depth}. While depth prediction from RGB images alone may be challenging, sparse depth measurements, e.g., obtained from keypoint triangulation, simplify the problem and lead to improvements in efficiency and accuracy \cite{Chen2018Depth}. Depth completion is the task of predicting dense depth images from sparse depth measurements and corresponding RGB images. Ma et al.~\cite{Ma2018Sparse,Ma2019SelfSupervised} develop methods for supervised training, relying on ground truth depth images, as well as self-supervised training, using photometric error from calibrated image pairs. Chen et al.~\cite{Chen2018Depth} pre-process sparse depth images by generating a Euclidean Distance Transform of the depth sample locations and a nearest-neighbor depth fill map. The authors propose a multi-scale deep network that treats depth completion as residual prediction with respect to the nearest-neighbor depth fill maps. 
Chen et al.~\cite{Chen2019Joint} design a 2-D convolution branch to process stacked RGB and sparse depth images and a 3-D convolution branch to process point clouds and fuse the outputs of the two branches.

\noindent \textbf{Mesh Reconstruction.}
Online terrain mapping requires efficient storage and updates of a 3-D surface model. Storing dense depth information from aerial images requires significant memory and subsequent model reconstruction. Explicit surface representations, e.g., based on polygon meshes, may be quite memory and computationally efficient but their vertices and faces need to be optimized to fit the environment geometry. FLaME \cite{Greene2017FlaME} performs non-local variational optimization over a time-varying Delaunay graph to obtain a real-time inverse-depth mesh of the environment. 
Terrian Fusion \cite{Wang2019TerrainFusion} performs real-time terrain mapping by generating digital surface model (DSM) meshes at selected keyframes. It converts the local meshes into grid-maps and merges them using multi-band fusion. 
Pixel2Mesh \cite{Wang2018Pixel2Mesh} treats a mesh as a graph and applies graph convolution \cite{kipf2017semi} for vertex feature extraction and graph unpooling to subdivide the mesh for detailed refinement. Mesh R-CNN \cite{Gkioxari2019MeshRCNN} simultaneously detects objects and reconstructs their 3-D mesh shape. A coarse voxel representation is predicted first and then converted into a mesh for refinement.

Our setting differs from existing work because aerial images cover large regions with significantly more subtle depth variation compared to indoor or outdoor ground settings. Our approach uses the same inputs as a depth completion problem but recovers a 3-D mesh model, which provides smoother depth estimates with fewer parameters (only vertices and faces) compared to dense depth prediction. Instead of relying on a priori known object categories, our method provides whole image mesh reconstruction.


\section{Problem Formulation}
\label{sec:problem_formulation}


Consider a UAV equipped with a camera, whose position $\bfp_k \in \mathbb{R}^3$ and orientation $\bfR_k \in SO(3)$ are estimated at discrete time steps $k$ by a VIO or SLAM localization algorithm. Denote the RGB images corresponding to the discrete-time keyframes by $\bfI_k$. Obtaining dense depth images during outdoor flight is challenging but it is common for localization algorithms to track and estimate the depth of a sparse set of image feature points. Let $\bfD_k^*$ denote the \emph{dense} ground-truth depth image, which is unknown during real-time operation. Let $\bfD_k$ be a \emph{sparse} matrix that contains estimated depths at the image feature locations and zeros everywhere else. Our goal is to construct an explicit model of the camera view at time $k$ using a 3-D triangle mesh $\calM_k := (\bfV_k,\calE_k,\calF_k)$, where $\bfV_k \in \mathbb{R}^{n_k \times 3}$ are the vertex coordinates in the camera frame, $[n_k] := \crl{1,\ldots,n_k}$ is the set of vertex indices, $\calE_k \subseteq [n_k] \times [n_k]$ are the edges, and $\calF_k \subseteq [n_k] \times [n_k] \times [n_k]$ are the faces.

\begin{problem*}
Given a finite set of RGB images $\crl{\bfI_k}_k$ and corresponding sparse depth measurements $\crl{\bfD_k}_k$, define a mesh reconstruction function $\calM = f(\bfI, \bfD ; \bftheta)$ and optimize its parameters $\bftheta$ to estimate the ground-truth depth $\crl{\bfD_k^*}_k$:
\begin{equation}
\min_{\bftheta} \sum_k \ell( f(\bfI_k, \bfD_k ; \bftheta), \bfD_k^* )
\end{equation}
where $\ell(\calM,\bfD)$ is a loss function measuring the error between a mesh $\calM$ and a depth image $\bfD$ representing the same camera view. 
\end{problem*}

The choice of loss function $\ell$ is discussed in Sec. \ref{sec:loss}. We develop a machine learning approach to this problem, consisting of an offline training phase and an online mesh reconstruction phase. During training, the parameters $\bftheta$ are optimized using a training set $\calD := \crl{ \bfI_i, \bfD_i, \bfD_i^* }_i$ with known ground-truth depth images. During testing, given streaming RGB images $\bfI_k$ and sparse depth measurements $\bfD_k$, the optimized parameters $\bftheta^*$ are used to construct mesh models $\calM_k = f(\bfI_k, \bfD_k; \bftheta^*)$. The local mesh $\calM_k = (\bfV_k,\calE_k,\calF_k)$ can be converted into the global frame by transforming the vertex coordinates $\bfV_k \bfR_k^\top + \mathbf{1} \bfp_k^\top$ using the camera poses $\bfp_k$, $\bfR_k$ and multiple meshes can be assembled \cite{Yu2004Poisson} to model the whole environment.



\section{Loss Functions for Mesh Reconstruction}
\label{sec:loss}

\begin{figure}[t]
  \centering
  \includegraphics[width=\linewidth,trim=0mm 8mm 0mm 9mm, clip]{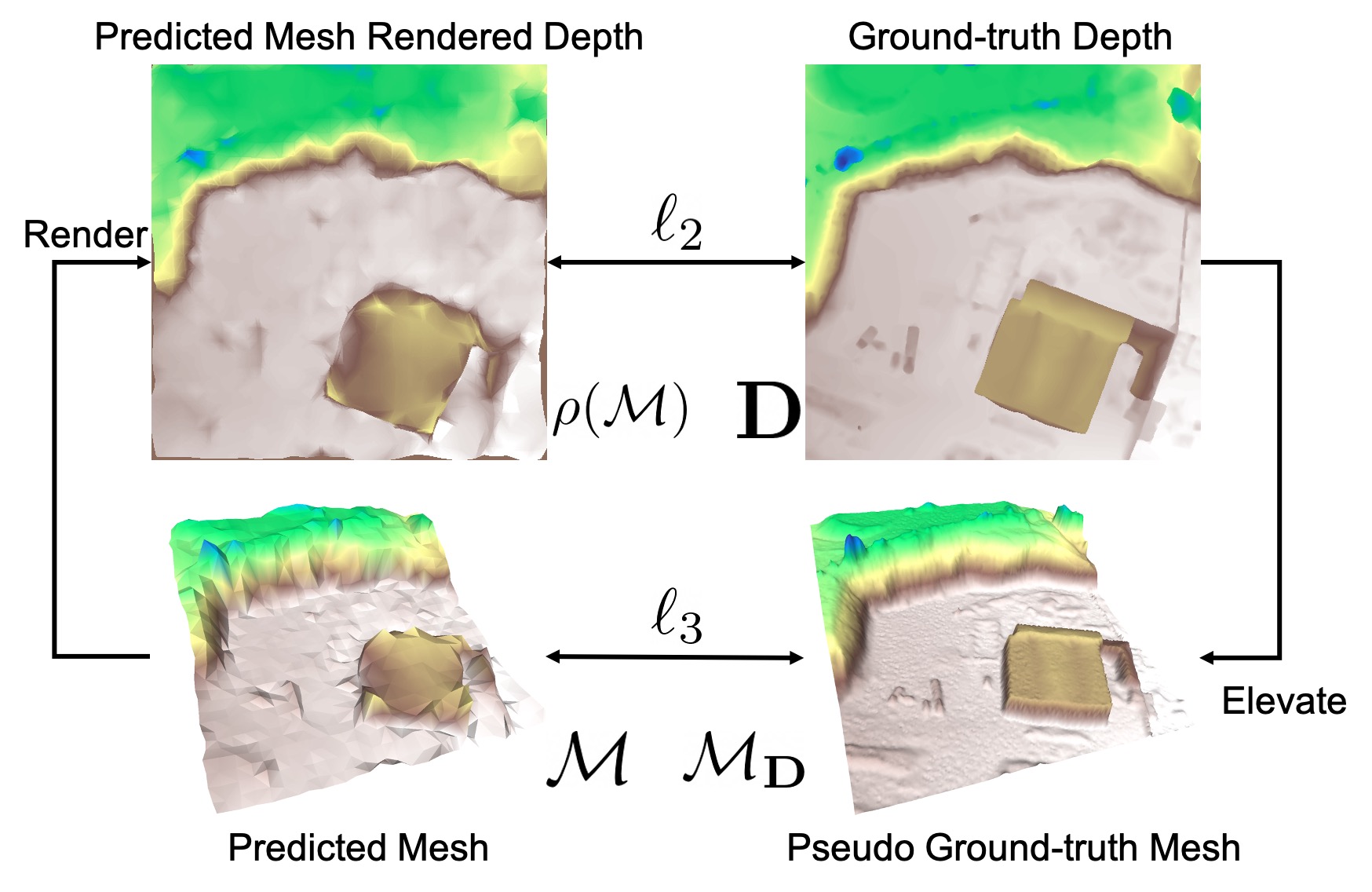}
  \caption{Loss function illustration: $\ell_{2}$ compares a depth image $\bfD$ to rendered mesh depth $\rho(\calM)$; $\ell_{3}$ compares a mesh $\calM$ to an elevated mesh $\calM_{\bfD}$ obtained from a depth image $\calD$.}
  \label{fig:supervision}
\end{figure}

We propose several loss functions to measure the error between a mesh $\calM$ and a depth image $\bfD$ representing the same camera view. Since our problem focuses on optimizing the mesh representation $\calM$, the loss function must be differentiable with respect to the vertices of $\calM$.

A loss function can be defined in the 2-D image plane by rendering a depth image from $\calM$ and comparing its pixel values to those of $\bfD$. We rely on a differentiable mesh renderer \cite{Liu2020Softras,ravi2020pytorch3d} to generate a depth image $\rho(\calM)$ and define a 2-D loss function:
\begin{equation}
  \label{eq:loss_2D}
  \ell_{2}(\mathcal{M},\bfD) := \frac{\sum_{ij\in\mathcal{U}(\rho(\mathcal{M}),\bfD)}|\rho_{ij}(\mathcal{M}) - \bfD_{ij}|}{|\mathcal{U}(\rho(\mathcal{M}),\bfD)|},
\end{equation}
where $\mathcal{U}(\rho(\mathcal{M}),\bfD)$ is the set of pixels where both the depth image $\bfD$ and the rendered depth $\rho(\mathcal{M})$ have valid depth information. While $\ell_2$ is a natural choice of a loss function in the image plane, it does not emphasize two important properties for mesh reconstruction. First, since $\ell_2$ only considers a region in the image plane where both depth images have valid information, its minimization over $\calM$ may encourage the mesh $\calM$ to shrink.
Second, $\ell_2$ does not emphasize regions of large depth gradient variation (e.g., the side surface of a building), which may lead to inaccurate reconstruction of 3-D structures.

To address these limitations, we propose a supplementary loss function, defined in the 3-D spatial domain over point clouds $\calP_{\calM}$ and $\calQ_{\bfD}$ obtained from $\calM$ and $\bfD$, respectively:
\begin{equation}
  \label{eq:loss_3D}
  \ell_{3}(\mathcal{M},\bfD) := \frac{1}{2} \lambda(\calP_{\calM},\calQ_{\bfD}) + \frac{1}{2}\lambda(\calQ_{\bfD},\calP_{\calM}),
\end{equation}
where $\lambda$ is the asymmetric Chamfer point cloud distance:
\begin{equation}
\lambda( \calP, \calQ) := \frac{1}{|\calP|} \sum_{\bfp \in \calP} \| \bfp - \argmin_{\bfq \in \calQ} \|\bfp - \bfq\|_2 \|_2.
\end{equation}
To generate $\calP_{\calM}$, we sample the faces of $\calM$ uniformly. Since samples on the mesh surface can be represented using barycentric coordinates with respect to the mesh vertices, the loss function is differentiable with respect to the mesh vertices. To generate $\calQ_{\bfD}$, we may sample the depth image $\bfD$ uniformly and project the samples to 3-D space but this will not generate sufficient samples in the regions of large depth gradient variation. Instead, we first generate a pseudo ground-truth mesh $\calM_{\bfD}$ by densely sampling pixel locations in $\bfD$ as the mesh vertices and triangulating on the image plane to generate faces. We then sample the surface of $\calM_{\bfD}$ uniformly to obtain $\calQ_{\bfD}$. 

Fig.~\ref{fig:supervision} illustrates the loss functions $\ell_2$ and $\ell_3$. We also define two regularization terms to measure the smoothness of $\calM = (\bfV, \calE, \calF)$. The first is based on the Laplacian matrix $\bfL := \bfG - \bfA \in \mathbb{R}^{n \times n}$ of $\calM$, where $\bfG$ is the vertex degree matrix and $\bfA$ is the adjacency matrix. We define a vertex regularization term based on the $\ell_{2,1}$-norm~\cite{Nie2010efficient} of the degree-normalized Laplacian \cite{Sorkine2004Laplacian} $\bfG^{-1}\bfL$:
\begin{equation}
\label{eq:loss_laplacian}
\ell_{\bfV}(\calM) := \frac{1}{n} \left\|\bfG^{-1}\bfL\bfV\right\|_{2,1}.
\end{equation}
We also introduce a mesh edge regularization term to discourage long edges in the mesh
\begin{equation}
\label{eq:loss_edge}
\ell_{\calE}(\calM) := \frac{1}{|\calE|} \sum_{(i,j) \in \calE} \|\bfv_i - \bfv_j\|_2,
\end{equation}
where $\bfv_i\in\mathbb{R}^3$ are the mesh vertices. The complete loss function is:
\begin{equation}
\label{eq:final_loss}
\begin{aligned}
\ell(\calM, \bfD) := w_2 &\ell_2(\calM, \bfD) + w_3 \ell_3(\calM, \bfD)\\
& + w_{\bfV} \ell_{\bfV}(\calM) + w_{\calE} \ell_{\calE}(\calM),
\end{aligned}
\end{equation}
where the first two terms evaluate the error between $\calM$ and $\bfD$ and the last two terms encourage smoothness of the mesh structure. The scalars $w_2, w_3, w_{\bfV}, w_{\calE}$ allow appropriate weighting of the different terms in \eqref{eq:final_loss}.


\section{2D-3D Learning for Mesh Reconstruction}
\label{sec:technical_approach}

\begin{figure*}[t]
  \centering
  \includegraphics[width=\linewidth,trim=0mm 15mm 0mm 10mm, clip]{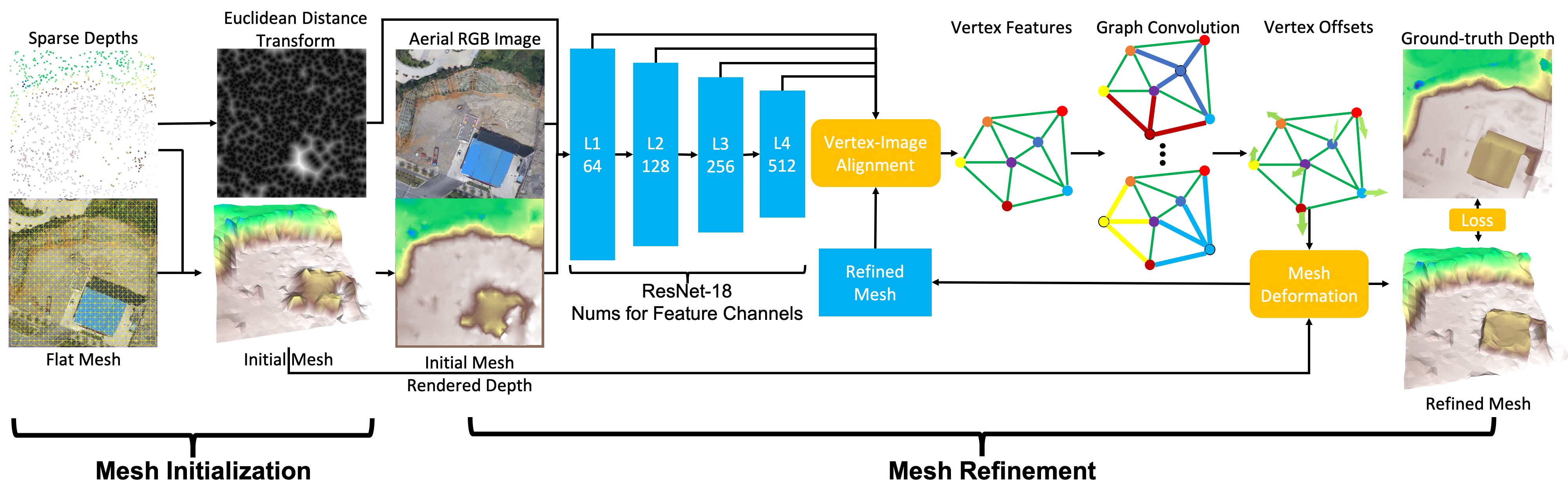}
  \caption{Overview of the complete mesh reconstruction architecture. In the mesh initialization stage (Sec.~\ref{sec:mesh_init}), we use sparse depths to elevate a flat mesh from the image plane to 3-D space. In the mesh refinement stage (Sec.~\ref{sec:mesh_refine}), we combine the RGB image, the initial mesh rendered depth and the Euclidean Distance Transform from the sparse depths and extract 2-D features using ResNet-18~\cite{He2016resnet}. These features are aligned to the initial mesh vertices using camera projection (Fig.~\ref{fig:vertex2image}). Vertex offsets are regressed using a graph convolution network (GCN) over the initial mesh. The ResNet-18 and GCN parameters are optimized jointly using the loss function in Sec.~\ref{sec:loss}.}
  \label{fig:pipeline}
\end{figure*}


Inspired by depth completion techniques, we approach mesh reconstruction in two stages: \emph{initialization} and \emph{refinement}. We initialize a coarse mesh from the sparse depth measurements and refine it by predicting vertex residuals based on RGB image features.

\subsection{Mesh Initialization}
\label{sec:mesh_init}
Outdoor terrain structure can be viewed as a 2.5-D surface that is mostly flat with occasional height variations. Hence, we initialize a flat mesh and change the surface elevation based on the sparse depth measurements. The flat mesh is initialized with regular-grid vertices (1024 in our experiments) over the X-Y ground plane, orthogonal to the gravity direction (Z axis). See Fig.~\ref{fig:pipeline} for an illustration. Subsequently, our mesh reconstruction approach only optimizes the mesh vertices and keeps the edge and face topology fixed. We initialize the Z-axis vertex coordinates by solving an optimization problem with a weighted combination of the 2-D rendered depth loss $\ell_2$ in \eqref{eq:loss_2D} and the Laplacian loss $\ell_\bfV$ in \eqref{eq:loss_laplacian} as the objective function:
\begin{equation}
\begin{aligned}
    \Delta \bfV^* = \argmin_{\Delta \bfV} \;\;&w_2\ell_{2}(\mathcal{M}(\bfV+\Delta \bfV,\calE, \calF),\bfD) \\
    &+ w_{\bfV}\ell_{\bfV}(\mathcal{M}(\bfV+\Delta \bfV,\calE,\calF)).
\end{aligned}
\label{eq:depth_opt}
\end{equation}
In \eqref{eq:depth_opt}, $\bfD$ are sparse depth measurements so the rendered depth error is evaluated only at the sparse pixel locations. The initialized mesh $\mathcal{M}^{\text{int}} = (\bfV+\Delta \bfV ^*,\calE, \calF)$ is used as an input to the mesh refinement stage.



\subsection{Mesh Refinement}
\label{sec:mesh_refine}

\begin{figure}[t]
  \centering
  \includegraphics[width=\linewidth,trim=0mm 8mm 0mm 5mm, clip]{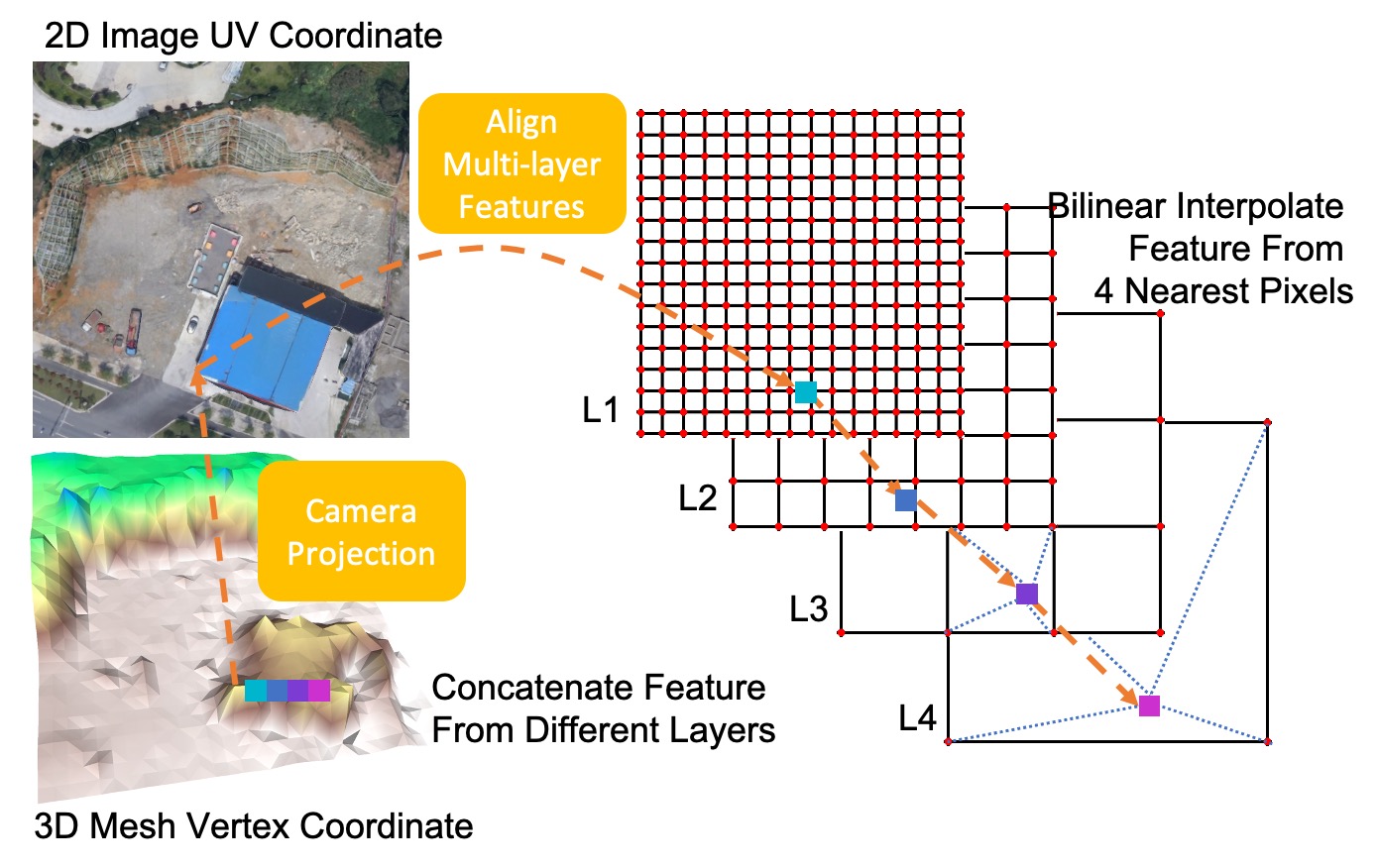}
  \caption{Illustration of image feature to mesh vertex association. With known camera intrinsics, each mesh vertex can be projected in uv coordinates (range $[0,1]$) onto the image plane. Bilinear interpolation is used to associate image feature maps at different resolutions with the mesh vertices. The features across different resolutions are concatenated to form a composite vertex feature.}
  \label{fig:vertex2image}
\end{figure}

In the refinement stage, we use a learning approach to extract features from both the 2-D image and 3-D initial mesh and regress mesh vertex offsets. The ground-truth depth maps are used for supervision.

The RGB image provides useful information for refinement since man-made objects have sharp vertical surfaces, while natural terrain has noisy but limited depth variation. The sparse depth measurements also provide information about areas with large intensity variation. Inspired by Mesh R-CNN~\cite{Gkioxari2019MeshRCNN}, we design a network that extracts features from the 2-D image, associates them with the 3-D vertices of the initial mesh, and uses them to refine the vertex locations. Our network has 3 stages: feature extraction, vertex-image feature alignment, and vertex graph convolution.



\noindent \textbf{Feature Extraction.}
We extract features from three sources: the RGB image $\bfI$, the rendered depth $\rho(\mathcal{M}^{\text{int}})$ from the initial mesh, and a Euclidean distance transform $\bfE(\bfD)$ of the sparse depth measurements $\bfD$, obtained by computing the Euclidean distance to the closest valid depth measurement for each pixel. The three images are concatenated to form a 5-channel input: $\bfC = \text{concat}(\bfI, \rho(\mathcal{M}^{\text{int}}), \bfE(\bfD))$. We use ResNet-18~\cite{He2016resnet} to extract image features. Since aerial images have different properties compared to ImageNet data, we learn the model weights from scratch. Four layers of features with different resolution and channels are extracted:
\begin{equation}
    [\bfL_1,\bfL_2,\bfL_3,\bfL_4] = \phi(\bfC),
    \label{eq:feature_2D}
\end{equation}
where $\phi$ is the ResNet-18 model.


\noindent \textbf{Vertex-Image Feature Alignment.} 
Next, we construct 3-D features for $\mathcal{M}^{\text{int}}$ by associating the mesh vertices with the 2-D image features. This idea is inspired by Pixel2Mesh \cite{Wang2018Pixel2Mesh}, which projects mesh vertices onto the image plane and extracts features at the projected coordinates. To obtain multi-scale features, we associate the projected mesh vertices with the intermediate layer feature maps $[\bfL_1,\bfL_2,\bfL_3,\bfL_4]$ from \eqref{eq:feature_2D}. This vertex-image alignment step is illustrated in Fig.~\ref{fig:vertex2image}. All features with different channel numbers are concatenated to form composite vertex features:
\begin{equation}
    \bfV^{g_{\text{in}}} = g^{\text{align}}(\mathcal{M}^{\text{int}},\phi(\bfC)),
    \label{eq:vertex_align}
\end{equation}
where $\bfV^{g_{\text{in}}} \in \mathbb{R}^{n\times(l_1+l_2+l_3+l_4+3)}$ are the vertex features and $l_i$ is the number of channels in feature map $\bfL_i$. We add the 3-D vertex coordinates $\bfv_i$ as the last 3 dimensions.


\noindent \textbf{Vertex Graph Convolution.} 
The mesh can be viewed as a graph with vertex features $\bfV^{g_{\text{in}}}$. A graph convolution network \cite{kipf2017semi,Gkioxari2019MeshRCNN} is a suitable architecture to process the vertex features and obtain vertex offsets $\Delta \bfV$ that optimize the agreement of the refined mesh $\mathcal{M}^{\text{ref}} = (\bfV^{\text{int}}+ \Delta \bfV,\calE, \calF)$ and the ground truth depth $\bfD^*$ according to the loss in \eqref{eq:final_loss}. To capture a larger region of feature influence, we use 3 layers of graph convolution $g_1,g_2,g_3$ and set the final vertex feature dimension to $64$. A weight matrix $\bfW$ is applied on the final $64$-D feature to derive the 3-D vertex offsets:
\begin{equation}
    \Delta \bfV = \bfW \bfV^{g_{\text{out}}} := \bfW g_3(g_2(g_1(\bfV^{g_{\text{in}}}))),
    \label{eq:vertex_offset}
\end{equation}
where $\bfV^{g_{\text{out}}} \in \mathbb{R}^{n\times64}$ and $\Delta \bfV \in \mathbb{R}^{n\times3}$. In order to refine the mesh with more details, we concatenate 3 stages of vertex-image feature alignment and graph convolution. At stage $i$, last stage's refined mesh $\mathcal{M}^{\text{ref}}_{i-1}$ is set as the initial mesh $\mathcal{M}^{\text{int}}_i$ and new vertex features are extracted via vertex-image feature alignment and fed to new graph convolution layers. All 3 refined meshes in different stages $(\mathcal{M}^{\text{ref}}_1,\mathcal{M}^{\text{ref}}_2,\mathcal{M}^{\text{ref}}_3)$ are evaluated against the ground-truth depth map $\bfD^*$ using the loss functions defined in \eqref{eq:final_loss}.

\section{Experiments}
\label{sec:experiments}

\begin{figure}[t]
  \centering
  \includegraphics[width=\linewidth,trim=0mm 0mm 0mm 0mm, clip]{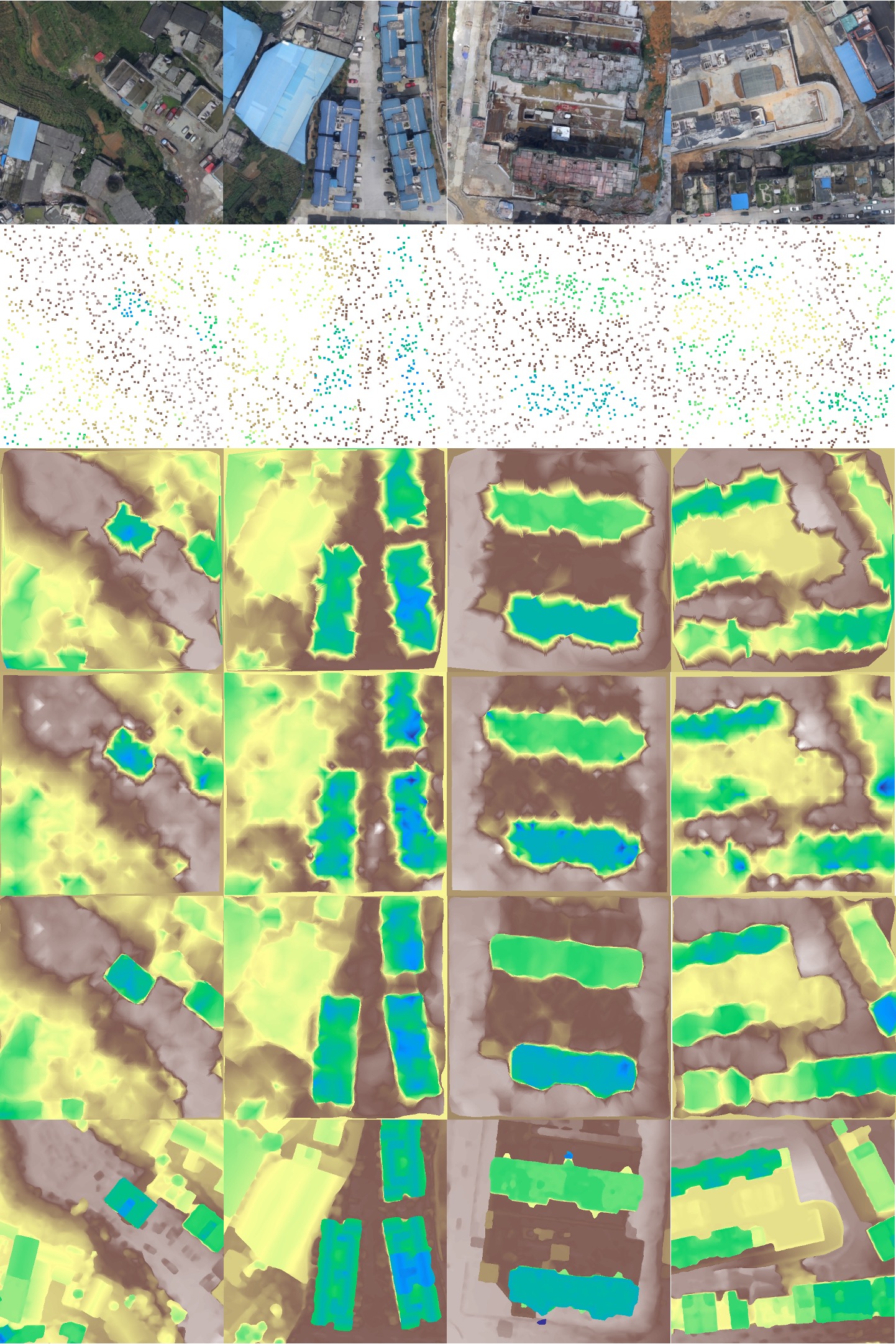}
  \caption{Mesh reconstructions visualized as rendered depth images. The colors indicate the relative depth values. Row 1: RGB images. Row 2: sparse depth measurements (around 1000). Row 3: meshes reconstructed from sparse-depth triangulation. Row 4: meshes after initialization (Sec. \ref{sec:mesh_init}). Row 5: meshes after neural network refinement (Sec. \ref{sec:mesh_refine}). Row 6: ground-truth depth images.}
  \label{fig:qual_results}
\end{figure}

\begin{figure}[t]
  \centering
  \includegraphics[width=\linewidth,trim=0mm 10mm 0mm 10mm, clip]{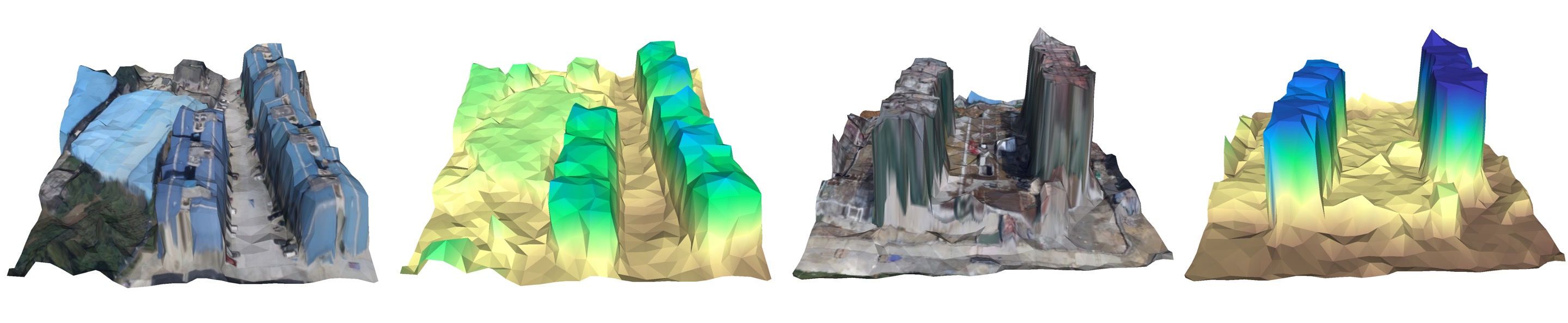}
  \caption{Reconstructed meshes painted with RGB texture and color indicating depth. The sharp vertical transitions of the buildings are reconstructed accurately.}
  \label{fig:qual_3d_results}
\end{figure}

\begin{figure}[t]
  \centering
  \includegraphics[width=\linewidth,trim=0mm 20mm 0mm 10mm, clip]{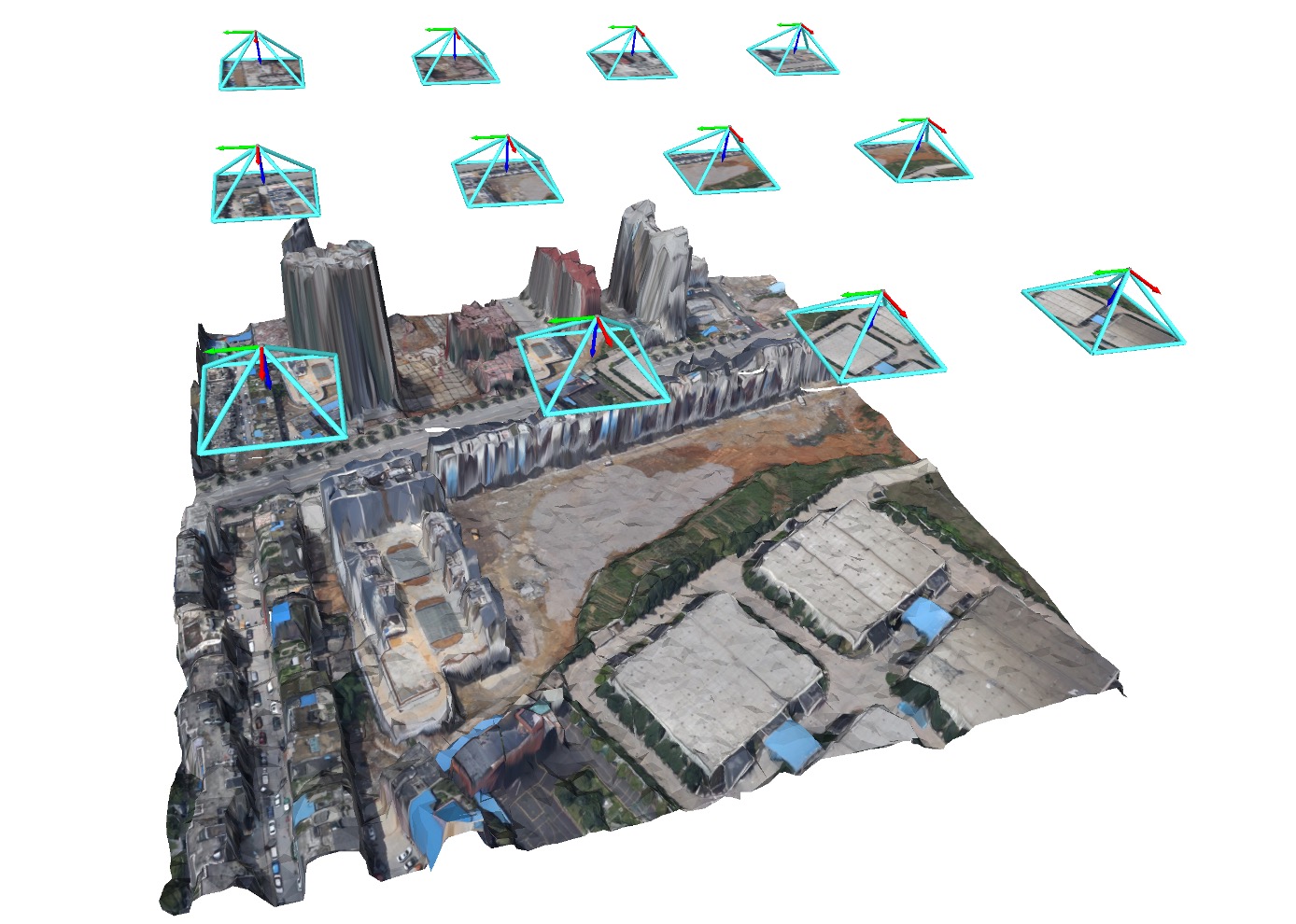}
  \caption{Complete environment model obtained by transforming to the global frame and merging local meshes from 12 camera views.}
  \label{fig:global_map}
\end{figure}


\begin{table*}[t]
  \caption{Quantitative evaluation of variations of the proposed method. The \emph{SD-tri} method triangulates a mesh using all the sparse depth measurements as vertices. The Regular-$n$ uses the mesh with $n$ vertices. The \emph{Initialized} model constructs a mesh from the sparse depth (Sec.~\ref{sec:mesh_init}). The RGB, RGB+RD, RGB+RD+EDT methods refine the initialized mesh (Sec.~\ref{sec:mesh_refine}), using different inputs respectively. The loss function used by the different methods is indicated by 3D+2D (uses both $\ell_3$ and $\ell_2$) or 3D (uses $\ell_3$ only). The second column shows the number of available sparse depth measurements per image and indicates whether the measurements are noisy (Sec.~\ref{sec:dataset}).}
  \label{tab:quantitative}
  \centering
  \begin{tabular}{|c|c|c|ccc|ccccc|}
        \hline
        \multirow{3}{*}{Error} & Meshing & SD-tri & \multicolumn{3}{c|}{Regular-576} & \multicolumn{5}{c|}{Regular-1024} \\
        \cline{2-11}
        & Inputs & \multirow{2}{*}{\shortstack{(vert \\= SD)}} & \multirow{2}{*}{Initialized} & RGB+RD & RGB+RD+EDT & \multirow{2}{*}{Initialized} & RGB & RGB+RD & RGB+RD & RGB+RD+EDT \\
        & Loss & & & 3D+2D & 3D+2D & & 3D+2D & 3D & 3D+2D & 3D+2D \\
        \hline 
        \multirow{3}{*}{$\ell_2$} & 500 & 1.492 & 2.069 & 1.670 & \textbf{1.637} & 1.861 & 1.575 & 1.382 & 1.289 & \textbf{1.252}\\
        & 1000 & 1.172 & 1.834 & 1.596 & \textbf{1.546} & 1.535 & 1.298 & 1.193 & 1.124 & \textbf{1.097}\\
        & 2000 & 0.916 & 1.941 & 1.551 & \textbf{1.511} & 1.344 & 1.144 & 1.092 & 1.045 & \textbf{1.024}\\
        \hline
        \multirow{3}{*}{$\ell_3$} & 500 & 9.815 & 18.278 & \textbf{13.438} & 13.763 & 13.799 & 7.242 & \textbf{5.412} & 5.647 & 6.352\\
        & 1000 & 6.494 & 17.762 & \textbf{12.938} & 13.574 & 11.872 & 5.876 & \textbf{4.538} & 4.911 & 5.703\\
        & 2000 & 4.649 & 17.130 & \textbf{12.483} & 13.506 & 10.859 & 5.131 & \textbf{4.069} & 4.477 & 5.291\\
        \hline 
        \hline
        \multirow{3}{*}{$\ell_2$} & 500+noise & 1.865 & 2.294 & 1.809 & \textbf{1.768} & 2.155 & 1.828 & 1.571 & 1.486 & \textbf{1.456}\\
        & 1000+noise & 1.632 & 2.056 & 1.701 & \textbf{1.685} & 1.826 & 1.535 & 1.360 & 1.319 & \textbf{1.308}\\
        & 2000+noise & 1.485 & 1.717 & 1.655 & \textbf{1.654} & 1.629 & 1.364 & 1.243 & \textbf{1.236} & 1.241\\
        \hline
        \multirow{3}{*}{$\ell_3$} & 500+noise & 19.737 & 18.392 & \textbf{12.974} & 13.532 & 14.887 & 8.351 & \textbf{6.063} & 6.157 & 6.865\\
        & 1000+noise & 22.189 & 17.693 & \textbf{12.258} & 13.161 & 12.480 & 6.447 & \textbf{4.904} & 5.266 & 6.075\\
        & 2000+noise & 18.545 & 17.256 & \textbf{11.856} & 12.988 & 11.147 & 5.452 & \textbf{4.343} & 4.793 & 5.620\\
        \hline
  \end{tabular}
\end{table*}

This section compares several variations of our mesh reconstruction approach to a baseline method on a dataset generated from aerial images.

\subsection{Dataset}
\label{sec:dataset}

We build an aerial image dataset based on the WHU MVS/Stereo dataset~\cite{Liu2020WHU}. The original dataset provides calibrated RGBD images rendered from a highly accurate 3D digital surface model which is not publicly available.
Hence, we recover a dense point cloud from the RGBD images as a ground-truth 3D model. We generate 20 camera trajectory sequences, split into 14 for training, 2 for validation, and 4 for testing. Each camera trajectory follows a sweeping grid-pattern with 10 keyframes per row and 20 keyframes per column. The keyframes are chosen to ensure 75\% row overlap and 80\% column overlap. RGBD images with resolution 512$\times$512 are rendered along each trajectory from the ground-truth point cloud using PyTorch3D~\cite{ravi2020pytorch3d}. We obtain camera pose estimates and sparse depth measurements $\bfD_k$ for each image by applying OpenSfM~\cite{OpenSfM} to four neighbor images with known camera intrinsic parameters. Small (500), medium (1000), and large (2000) number of sparse depth measurements are obtained from SfM per image. The results from OpenSfM are treated as the data with noise, and the noise may come from the feature detection\&matching as well as the bundle adjustment step. We also obtain noiseless depth measurements with the same sparsity 2-D pattern from the ground-truth depth images $\bfD_k^*$.


\subsection{Implementation Details}
\label{sec:impl_details}

During training, we use 1000 number of sparse depth measurements and the mesh vertices number is fixed to 1024. The mesh initialization optimization is performed over 100 iterations with the Adam optimizer~\cite{kingma2014adam} and weights $[w_2, w_3, w_{\bfV}, w_{\calE}] = [1,0,0.5,0]$ for the loss function in \eqref{eq:final_loss}. The ResNet-18 and GCN parameters are optimized jointly during the mesh refinement training using the Adam optimizer with initial learning rate of $0.0005$ for $200$ epochs. The loss function in \eqref{eq:final_loss} with parameters $[w_2, w_3, w_{\bfV}, w_{\calE}] = [3,1,0.5,0.01]$ is used in this phase. The Chamfer distance $\lambda$ in the $\ell_3$ loss term is computed using 10000 samples.


\subsection{Results}

Our experiments report the $\ell_2$ error in \eqref{eq:loss_2D} and the $\ell_3$ error in \eqref{eq:loss_3D}, comparing the reconstructed mesh wit. The $\ell_2$ emphasizes projected depth accuracy, while the $\ell_3$ pay more attention to large depth gradient region.

For comparison, we define a baseline method that triangulates the sparse depth measurements directly to build a mesh. The baseline method performs Delaunay triangulation on the 2-D image plane over the depth measurements and projects the flat mesh to 3-D using the known vertex depths. We refer to the baseline method as sparse-depth-triangulation (\emph{SD-tri}). Note that the baseline method uses all avaiable sparse depth measurements (500, 1000, or 2000) and, hence, may has a different number of vertices from the other models. The quantitative results from the comparison are reported in Table \ref{tab:quantitative}. Note that all the models are trained with 1024-vertex meshes and 1000 sparse depth measurements and we directly generalize them on meshes with different number of vertices and different number of sparse depth measurements.

Several variations of our approach are evaluated. We compared three different options for the input provided to the mesh refinement stage: only the 3-channel RGB image (RGB), the RGB image plus rendered depth from the initial mesh (RGB+RD, 4-channels), and the RGB image plus rendered depth from the initial mesh plus Euclidean distance transform obtained from of the sparse depth map (RGB+RD+EDT, 5-channels). The model using RGB-only does not perform as well as the other two. The RGB+RD+EDT model has the best performance according to the $\ell_2$ error metric. The RGB+RD method has similar performance in the $\ell_2$ metric and smaller $\ell_3$ error compared to RGB+RD+EDT. The RGB+RD model is used to generate our qualitative results using 1024-vertex meshes because it offers good performance according to both error metrics.

We also compare different loss function combinations for training the RGB+RD method. The 3D+2D loss function reported in Table \ref{tab:quantitative} corresponds to training with parameters $[w_2, w_3, w_{\bfV}, w_{\calE}] = [3,1,0.5,0.01]$ for the loss function in \eqref{eq:final_loss}, while the 3D loss function, corresponds to parameters $[w_2, w_3, w_{\bfV}, w_{\calE}] = [0,1,0.5,0.01]$. We can see that training with the 3D+2D loss leads to balanced performance acorrding to both the $\ell_2$ and $\ell_3$ metrics, while training with the 3D loss only leads to good performance in the $\ell_3$ loss but higher 2-D rendering error, according to the $\ell_2$ metric.

Finally, we compare the mesh reconstruction accuracy when the sparse depth measurements are noiseless versus noisy. The baseline SD-tri method performs well in a noiseless setting but degenerates drastically when noise from the SfM is introduced. In contrast, our model is more robust to the noise in the sparse depth measurements. Two factors might be contributing to this. First, our mesh initialization and refinement stages both include explicit mesh regularization terms (in \eqref{eq:loss_laplacian} and \eqref{eq:loss_edge}). Second, the image features extracted during the mesh refinement process help to distinguish among different terrains and structures. The latter is clear from the improved accuracy of the refined, compared to the initialized, meshes. We also report the performance using a mesh with only 576 vertices. When the depth measurements are noisy, it has lower $\ell_2$ loss compared with the baseline method with similar number of vertices. It also has lower $\ell_3$ loss even compared with meshes with more vertices generated from the baseline method.

Some qualitative results are presented in Fig.~\ref{fig:qual_results}. Compared with the sparse-depth-triangulation and the initialized meshes, the refined meshes have smoother boundaries on the side surfaces of the buildings. The guidance from the image features allows the refined meshes to fit the 3D structure better. Fig.~\ref{fig:global_map} shows a global mesh reconstruction obtained by transforming and merging 12 local camera-view reconstructions. The local meshes are transformed to global frame using the keyframe poses and no postprocessing is used to merge them into a global model.

The reconstructed mesh models are a more efficient representation than the dense depth images. A dense depth image requires $512\times 512$ values to represent a camera view, while our mesh model (with fixed mesh faces topology) only needs to store the 3D coordinates of the 1024 vertices. Thus, our model requires only $1\%$ of the depth image parameters to obtain a high-fidelity reconstruction of a camera view. On a desktop with NVIDIA 1080 Ti GPU, the Mesh Initialization step takes about 3 s/frame because we solve it using iterative gradient descent. However, eq. \eqref{eq:depth_opt} can be solved much faster by treating it as a linear system. The Mesh Refinement step takes about 10 ms/frame.

\section{Conclusion}
\label{sec:conclusion}

In this work, we introduce a method to reconstruct a 3D mesh from an RGB image and sparse depth measurements. We build an outdoor aerial dataset and apply our method on it for terrain mapping. Quantitative and qualitative results show that our method outperform the baseline method of triangulating the sparse depth points. Our method can also generalize to different number of sparse depth without additional storage cost. It is also robust to reconstruction noise in the sparse depth measurements. In the future, we would like to tightly merge this with a feature-based SLAM algorithm to upgrade the sparse feature-based map into a dense mesh map. We would also fuse multi-modal observations into the map such as the semantic information.


\balance
{\small
\bibliographystyle{cls/IEEEtran}
\bibliography{bib/ref.bib}
}
\end{document}